\documentclass[11pt,letterpaper]{article}
\usepackage[margin=1in]{geometry}
\usepackage[utf8]{inputenc}
\usepackage[T1]{fontenc}
\usepackage{microtype}
\usepackage{graphicx}
\graphicspath{{./}{../figures/}}
\usepackage{booktabs}
\usepackage{tabularx}
\usepackage{caption}
\usepackage{float}
\usepackage{amsmath,amssymb,mathtools}
\usepackage{xcolor}
\usepackage[numbers,sort&compress]{natbib}
\usepackage{hyperref}
\usepackage{enumitem}
\usepackage{needspace}
\usepackage{placeins}
\usepackage{url}
\usepackage{tikz}
\usetikzlibrary{arrows.meta,positioning,calc}

\hypersetup{
  pdftitle={Agent Harness Awareness: Turning Verified Telemetry into Operational Control for Tool-Using Agents},
  pdfauthor={Manu Agrawal and Shrey Nagpal},
  pdfsubject={Host-verified operational telemetry for tool-using language agents},
  pdfkeywords={AI agents, tool use, operational telemetry, OpenTelemetry, reliability},
  colorlinks=true,
  linkcolor=blue!70!black,
  citecolor=blue!70!black,
  urlcolor=blue!70!black
}

\makeatletter
\renewcommand\section{\@startsection {section}{1}{\z@}%
                                   {-1.4ex \@plus -0.3ex \@minus -.2ex}%
                                   {0.5ex \@plus .2ex}%
                                   {\normalfont\Large\bfseries}}
\renewcommand\subsection{\@startsection{subsection}{2}{\z@}%
                                     {-1.1ex\@plus -0.2ex \@minus -.2ex}%
                                     {0.4ex \@plus .2ex}%
                                     {\normalfont\large\bfseries}}
\renewcommand{\@seccntformat}[1]{\csname the#1\endcsname\hspace{0.6em}}
\makeatother

\title{\textbf{Agent Harness Awareness: Turning Verified Telemetry into Operational Control for Tool-Using Agents}}
\author{
  Manu Agrawal \qquad Shrey Nagpal \\
  \small Independent Researchers
}
\date{}

\begin{document}

\maketitle

\begin{abstract}
Tool descriptions tell an agent what an action can do, but they do not necessarily reveal whether that action is healthy or suitable at the moment of use. Execution harnesses often possess this information through monitors, traces, and recent failures while the model choosing the next action does not. We call this gap \textbf{operational blindness}. We introduce \textbf{Agent Harness Awareness}, a trust-separated design in which the host supplies a small, fresh, verified state snapshot before a consequential action while retaining measurement, enforcement, evaluation, and final dispatch authority. We test its tools-and-network instance with \textbf{ToolRoute}, a controlled isolation benchmark that holds the task and candidate semantics constant and varies the operational information available to the model. In a 216-decision ablation across GPT-5.6 Sol, Claude Sonnet 5, and Gemini 3.7 Flash, verified tool-health state reduced mean observable policy regret by \textbf{96.9\%} against a same-shape neutral control (4,252\,ms to 131\,ms) and raised \textbf{single-dispatch completion} from \textbf{69.4\% to 95.8\%}. In a separate 27-decision end-to-end replication with live local TCP/HTTP execution, OpenTelemetry client spans, and Toxiproxy-injected faults, verified state reduced mean regret from \textbf{8,956\,ms to 0\,ms}---zero excess observable operational cost---and completed the live action in \textbf{9/9} model-by-fault cases versus \textbf{5/9} for the neutral control. These studies establish the tools-and-network instance of Agent Harness Awareness: selected host telemetry materially improves pre-dispatch operational choice. Runtime state, hardware pressure, economics, retry-aware recovery, and heterogeneous selection remain open empirical extensions of the same architecture.
\end{abstract}

\section{Introduction}

Consider an agent choosing between two actions that can both retrieve the required record. The tool descriptions and schemas say that either action is valid. The execution harness, however, has just observed repeated failures from one route and healthy responses from the other. If the model sees only the static descriptions, it cannot use what the harness already knows. It may produce a syntactically and semantically correct tool call that is nevertheless an avoidable operational error.

Tool-using architectures and benchmarks already treat external actions as central to agent behavior~\cite{yao2023react,schick2023toolformer,liu2024agentbench,yao2025taubench}. For the question studied here, it is useful to separate two decisions. \textbf{Semantic discovery} determines which actions can satisfy the task. \textbf{Operational selection} determines which eligible action is suitable under current conditions. ReliabilityBench evaluates 1,280 episodes under production-like perturbations and controlled tool/API faults, identifying rate limiting as its most damaging fault ablation~\cite{gupta2026reliabilitybench}; ToolMaze and ToolBench-X independently expose substantial diagnosis and recovery failures in unreliable tool environments~\cite{zhu2026whentoolsfail,tian2026toolbenchx}. Yet the operational evidence collected by monitors, traces, circuit breakers, and quota systems is commonly used by infrastructure or human operators without becoming an explicit input to the model's decision.

We call this information gap \textbf{operational blindness}. \textbf{Agent Harness Awareness} addresses it by giving the model a bounded, freshness-labeled, host-verified view of relevant execution state before a consequential action. The host remains authoritative: it measures and reduces the state, enforces constraints, revalidates the decision, and performs the action. The model can adapt its choice but cannot alter the telemetry or bypass enforcement.

ToolRoute isolates whether this information changes behavior. Its candidate routes are functionally equivalent by design: holding task semantics and candidate eligibility fixed reduces semantic, schema, and task-difficulty confounds, making the operational-information contrast directly measurable. This controlled probe tests a prerequisite for broader use: whether frontier models can convert verified current state into a better operational choice when capability is not in dispute.

The same mechanism could later support two more complex settings, neither evaluated here. A recovery-capable agent could use current state to choose among retrying, waiting, switching, or verifying after a failure. An agent selecting among heterogeneous eligible providers or data sources could weigh current availability, freshness, price, privacy, or quota against the requirements of the task. These are extensions of the tested decision mechanism, not results of this study.

This paper makes three contributions:
\begin{enumerate}[leftmargin=*,itemsep=0.5pt,parsep=0pt,topsep=1pt]
\item It defines a trust-separated interface for turning selected host observations into an agent decision input while preserving host authority over truth, safety, and dispatch.
\item It provides a controlled 216-decision information ablation across three frontier model families, including a same-shape neutral control that separates verified state values from interface structure.
\item It independently replicates the intervention across 27 decisions on a live local HTTP path instrumented with OpenTelemetry and Toxiproxy. The empirical claims in this preprint are limited to this tools-and-network instance.
\end{enumerate}

\section{Agent Harness Awareness: system architecture}

Agent Harness Awareness inserts one explicit information step between capability discovery and execution:
\begin{enumerate}[leftmargin=*,itemsep=0.5pt,parsep=0pt,topsep=1pt]
\item The host measures recent execution outcomes and current constraints.
\item A deterministic reducer selects relevant facts, bounds and redacts them, and attaches provenance and freshness.
\item The model chooses among semantically eligible actions using the task and this verified snapshot.
\item The host revalidates current state, enforces hard constraints, and dispatches or safely rejects the action.
\end{enumerate}

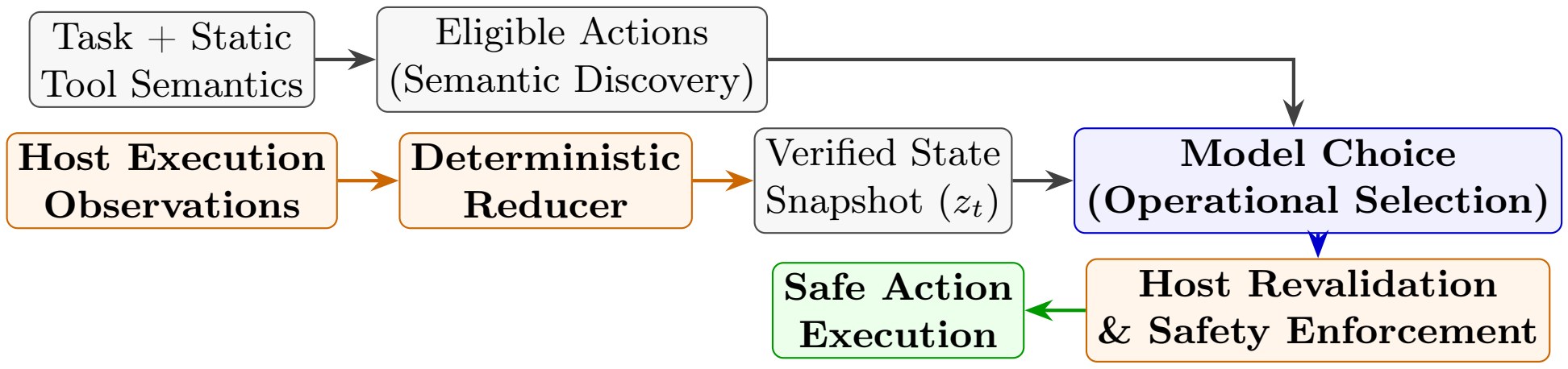
\begin{figure}[htbp]
\centering
\begin{tikzpicture}[
    >=Stealth,
    font=\footnotesize,
    box/.style={draw=black!70, rounded corners=2.5pt, fill=black!3, inner sep=2.5pt, align=center},
    host/.style={draw=orange!80!black, rounded corners=2.5pt, fill=orange!8, inner sep=2.5pt, align=center, font=\footnotesize\bfseries},
    agent/.style={draw=blue!80!black, rounded corners=2.5pt, fill=blue!6, inner sep=2.5pt, align=center, font=\footnotesize\bfseries},
    exec/.style={draw=green!60!black, rounded corners=2.5pt, fill=green!8, inner sep=2.5pt, align=center, font=\footnotesize\bfseries}
]
    \node[box] (task) {Task + Static\\Tool Semantics};
    \node[box, right=0.5cm of task] (eligible) {Eligible Actions\\(Semantic Discovery)};
    
    \node[host, below=0.20cm of task] (obs) {Host Execution\\Observations};
    \node[host, right=0.5cm of obs] (reducer) {Deterministic\\Reducer};
    \node[box, right=0.5cm of reducer] (state) {Verified State\\Snapshot ($z_t$)};
    
    \node[agent, right=0.5cm of state] (choice) {Model Choice\\(Operational Selection)};
    
    \node[host, below=0.20cm of choice] (enforce) {Host Revalidation\\\& Safety Enforcement};
    \node[exec, left=0.5cm of enforce] (exec) {Safe Action\\Execution};
    
    \draw[->, thick, black!75] (task) -- (eligible);
    \draw[->, thick, orange!80!black] (obs) -- (reducer);
    \draw[->, thick, orange!80!black] (reducer) -- (state);
    
    \draw[->, thick, black!75] (eligible.east) -| ($(choice.north) + (-0.2,0)$);
    \draw[->, thick, black!75] (state) -- (choice);
    
    \draw[->, thick, blue!80!black] (choice) -- (enforce);
    \draw[->, thick, green!60!black] (enforce) -- (exec);
\end{tikzpicture}
\vspace{-2pt}
\caption{Conceptual Agent Harness Awareness flow. Static capability information and reduced current state meet at the operational-selection boundary; measurement, validation, enforcement, and execution remain host-owned.}
\label{fig:conceptual_arch}
\end{figure}

At decision time $t$, let $x_t$ denote the task and eligible-action interface and $h_t$ the host operational state. The host constructs $z_t=\phi(h_t)$, a bounded projection containing selected facts relevant to the decision. The operator $\phi$ is not a learned policy or a transfer of safety authority: it denotes host-owned selection, reduction, redaction, serialization, and freshness labeling. The behavioral question is whether a policy $\pi(a_t\mid x_t,z_t)$ chooses a lower-loss operational action than a policy given $x_t$ alone or the same interface populated with neutral values.

We call this interface the \textbf{Substrate State Telemetry (SST) contract}. It organizes a broader research program while keeping the evidence boundary explicit:

\vspace{1pt}
{\small
\noindent\begin{tabularx}{\linewidth}{@{} l X l @{}}
\toprule
\textbf{SST dimension} & \textbf{Example host facts} & \textbf{Status in this paper} \\
\midrule
Tools and network & Recent success, latency, errors, circuit state, freshness & \textbf{Evaluated with ToolRoute} \\
Runtime & Deadline, queueing, concurrency, exit state & Prospective \\
Hardware and sandbox & Memory headroom, CPU pressure, storage & Prospective; evidence in~\cite{agrawal2026substrate} \\
Economics & Context headroom, tool-call budget, quota, price & Prospective; budget planning in~\cite{liu2025bats} \\
\bottomrule
\end{tabularx}
}
\vspace{1pt}

For ToolRoute, each route row contains a probe-window count, successes, latency EWMA, consecutive failures and last error, circuit state, and observation age. The snapshot header states its freshness bound. Routine replica failover should remain in deterministic middleware. Equivalent routes are used here as a controlled behavioral probe, not as a proposal to replace a proxy with an LLM. The motivating agent-level case arises when task semantics determine how operational state should be weighed---for example, whether to change provider, data source, execution mode, or plan---and requires further evaluation with heterogeneous candidates.

This work is adjacent to, but distinct from, several recent directions. Agent-Native Telemetry develops compact, verifiable telemetry representation, transport, and access paths~\cite{he2026agentnative}. BATS demonstrates budget-conditioned planning when remaining tool-call budget is made explicit~\cite{liu2025bats}. ReliabilityBench, ToolMaze, and ToolBench-X document reliability losses under tool/API faults and difficulties in diagnosis or recovery~\cite{gupta2026reliabilitybench,zhu2026whentoolsfail,tian2026toolbenchx}. ToolRoute asks a narrower controlled question: holding the task, candidates, and interface structure fixed, does truthful current host state change the consequential pre-dispatch action?

\section{Two complementary ToolRoute studies}

Study 1 estimates the behavioral effect of operational information. Study 2 separately tests whether the same intervention operates through an instrumented live HTTP path. Their evidence is never pooled.

\vspace{1pt}
{\small
\noindent\begin{tabularx}{\linewidth}{@{} >{\raggedright\arraybackslash}p{3.5cm} >{\raggedright\arraybackslash}X >{\raggedright\arraybackslash}X >{\raggedright\arraybackslash}p{2.8cm} @{}}
\toprule
\textbf{Study} & \textbf{Question} & \textbf{Evidence} & \textbf{Role in the paper} \\
\midrule
\textbf{Study 1: controlled information ablation} & Does truthful state improve route selection when the decision interface is held constant? & 216 frozen direct-provider decisions; seeded host-owned monitor. & Primary behavioral estimate. \\
\addlinespace[2pt]
\textbf{Study 2: live transport replication} & Does the same information effect survive a standard-telemetry, live HTTP tool path? & 27 frozen remote-model decisions; OTel spans, Toxiproxy, and live selected-route execution. & End-to-end systems replication; never pooled with Study 1. \\
\bottomrule
\end{tabularx}
}
\vspace{1pt}

\subsection{Study 1: controlled information ablation}

The task is to retrieve a read-only customer record through one of two functionally equivalent routes, \texttt{tool\_alpha} and \texttt{tool\_beta}, or choose \texttt{wait}. With task and candidate capability fixed, the experimental question is whether truthful current state changes the selected action. The host owns monitoring, schedules, execution, and an external observable-cost oracle. Each model receives an independent full-checkpoint episode and has no shell, filesystem, network-tool, schedule, oracle, or artifact access.

\vspace{1pt}
{\small
\noindent\begin{tabularx}{\linewidth}{@{} l X @{}}
\toprule
\textbf{Condition} & \textbf{Model-visible information} \\
\midrule
A: task only & Task and equivalent tool descriptions; no operational projection. \\
B: neutral envelope & Same telemetry-shaped envelope, fields, row order, and tool rows as C, with equal neutral values for both routes. \\
C: verified state & The same envelope with fresh host-owned facts for current route state. \\
\bottomrule
\end{tabularx}
}
\vspace{1pt}

\textbf{A is the deployment baseline:} task and tool semantics without harness state. \textbf{B is the structural control:} the same telemetry-shaped interface with route-relevant values neutralized. \textbf{C is the intervention:} verified current state. A-to-C therefore measures the practical single-dispatch difference from the task-only setup, while B-to-C supports attribution to route-relevant information rather than the presence or shape of the telemetry envelope. B is not an exact provider-token-matching control; input-token counts are retained descriptively. The literal neutral values are reported in Appendix~A.

We use two complementary outcomes. \textbf{Single-dispatch completion} records whether the selected route retrieves the required record on that attempt; selecting \texttt{wait} is a non-completion under the frozen task contract. \textbf{Observable policy regret} measures excess operational cost relative to the observable-best route under the same facts available in Condition~C. Thus 0\,ms regret means zero avoidable operational cost, not zero request duration. A route may complete yet have positive regret when it was avoidably slower or riskier.

Let $\mathcal{R}=\{\texttt{tool\_alpha},\texttt{tool\_beta}\}$. For a selected route $r\in\mathcal{R}$,
\begin{equation}
r_{\mathrm{obs}}(r,z)=c_{\mathrm{obs}}(r,z)-\min_{r'\in\mathcal{R}}c_{\mathrm{obs}}(r',z).
\end{equation}
For a route with monitor window $n$, $s$ successes, latency EWMA $\ell$, and circuit state $q$, let $p_f=(n-s)/n$ when $n>0$, and $p_f=0.5$ when $n=0$. The frozen route cost is
\begin{equation}
c_{\mathrm{obs}}(r,z)=\ell+10{,}000p_f+10{,}000\,\mathbf{1}[q=\mathrm{OPEN}]\ \text{ms}.
\end{equation}
The 10,000\,ms terms are frozen workflow-level failure and open-circuit penalties expressed in operational-cost units; they are not measurements of wire latency. The evaluator uses only canonical monitor facts rendered in C and never hidden latent anomalies. Appendix~A gives the separate \texttt{wait} rule, missing-window rule, and evaluated-schedule bound.

The frozen cohort contains 216 independent decisions: three model families $\times$ six seeds $\times$ four predeclared environment states per seed $\times$ three conditions. ``State'' denotes an independent checkpoint, not a conversational turn. No request was retried; this design measures the initial pre-dispatch choice before reactive recovery can obscure it. Every attempt retained its prompt, sanitized request and response, snapshot, tool event, evaluation result, and finalization hashes.

\subsection{Study 2: live OpenTelemetry transport replication}

The separately labeled v0.2 replication preserves the A/B/C intervention while replacing the seeded monitor with a live observation and execution path:

\begin{center}
\vspace{-6pt}
\begin{tikzpicture}[
    >=Stealth,
    font=\footnotesize,
    node distance=0.15cm and 0.25cm,
    box/.style={draw=black!70, rounded corners=2.5pt, fill=black!3, inner sep=2.2pt, align=center},
    otel/.style={draw=orange!80!black, rounded corners=2.5pt, fill=orange!8, inner sep=2.2pt, align=center, font=\footnotesize\bfseries},
    agent/.style={draw=blue!80!black, rounded corners=2.5pt, fill=blue!6, inner sep=2.2pt, align=center, font=\footnotesize\bfseries},
    exec/.style={draw=green!60!black, rounded corners=2.5pt, fill=green!8, inner sep=2.2pt, align=center, font=\footnotesize\bfseries}
]
    \node[box] (client) {Instrumented\\HTTP Client};
    \node[box, right=0.35cm of client] (proxy) {Toxiproxy\\Fault Injection};
    \node[box, right=0.35cm of proxy] (route) {Equivalent Local\\HTTP Route};
    
    \draw[->, thick, black!75] (client) -- (proxy);
    \draw[->, thick, black!75] (proxy) -- (route);
    
    \node[otel, below=0.12cm of proxy] (spans) {OpenTelemetry Client Spans};
    \draw[->, dashed, thick, orange!80!black] (client.south) |- (spans.west);
    \draw[->, dashed, thick, orange!80!black] (route.south) |- (spans.east);
    
    \node[box, below=0.16cm of spans] (state) {Bounded\\Agent State};
    \node[box, left=0.22cm of state] (reducer) {Host-Owned\\Reducer};
    \node[agent, right=0.22cm of state] (decision) {Model\\Decision};
    \node[exec, right=0.22cm of decision] (exec) {Live Selected-Route\\Execution};
    
    \draw[->, thick, orange!80!black] (spans.south) -| (reducer.north);
    \draw[->, thick, black!75] (reducer) -- (state);
    \draw[->, thick, black!75] (state) -- (decision);
    \draw[->, thick, blue!80!black] (decision) -- (exec);
\end{tikzpicture}
\vspace{-6pt}
\end{center}

Each episode sends three monitor requests per route through local TCP/HTTP proxies. OpenTelemetry HTTP client spans provide duration and outcome facts~\cite{opentelemetry2026spec}. The deterministic reducer converts the six spans into route-level counts, successes, latency EWMA ($\alpha=0.2$), error and circuit state, and a 2\,s freshness bound. A circuit is \texttt{OPEN} after three consecutive failed probes; otherwise a three-probe window with failure fraction at least 0.4 is \texttt{DEGRADED}. Toxiproxy injects a predeclared 300\,ms latency, connection failure, or HTTP 503 on one route; the HTTP client timeout is 2\,s~\cite{shopify2026toxiproxy}. The selected route is then executed through the same live proxy.

The replication contains 27 independent decisions: three model families $\times$ three fault regimes $\times$ A/B/C. Its manifest fixes the provider/model, condition, faulted route, option order, execution order, and no-retry policy before execution. Raw spans, reducer events, prompts, sanitized provider records, post-decision action spans, results, and finalization hashes are retained in the artifact archive. This cohort remains separate from Study 1.

\section{Results}

\subsection{Study 1: controlled information ablation}

The primary structural comparison is B-to-C. With the interface shape held constant, verified state reduces mean observable policy regret from 4,251.54\,ms to 131.12\,ms (\textbf{96.9\% lower}) and raises single-dispatch completion from 50/72 (69.4\%; Wilson 95\% CI 58.0--78.9\%) to 69/72 (95.8\%; Wilson 95\% CI 88.5--98.6\%), a \textbf{26.4 percentage-point increase}. A descriptive bootstrap over the 18 model-by-seed blocks gives a regret difference of 4,120\,ms (95\% interval 2,543--5,893\,ms) and a completion gain of 26.4 points (18.1--34.7 points).

The deployment comparison is consistent with that result. Relative to task-only A, C reduces mean regret from 4,826.45\,ms to 131.12\,ms (\textbf{97.3\% lower}) and raises completion from 47/72 (65.3\%; Wilson 95\% CI 53.8--75.2\%) to 69/72 (95.8\%), a \textbf{30.6 percentage-point increase}.

\begin{table}[htbp]
\centering
\small
\caption{Study 1 performance across information conditions.}
\label{tab:study1_results}
\begin{tabularx}{\linewidth}{@{} l X r r @{}}
\toprule
\textbf{Condition} & \textbf{Role} & \textbf{Mean regret (ms)} & \shortstack{\textbf{Single-dispatch}\\\textbf{completion}} \\
\midrule
A & Task-only deployment baseline & 4,826.45 & 47/72 (65.3\%) \\
B & Neutral structural control & 4,251.54 & 50/72 (69.4\%) \\
C & Verified host state & 131.12 & 69/72 (95.8\%) \\
\bottomrule
\end{tabularx}
\end{table}

\begin{figure}[htbp]
\centering
\includegraphics[width=0.93\textwidth]{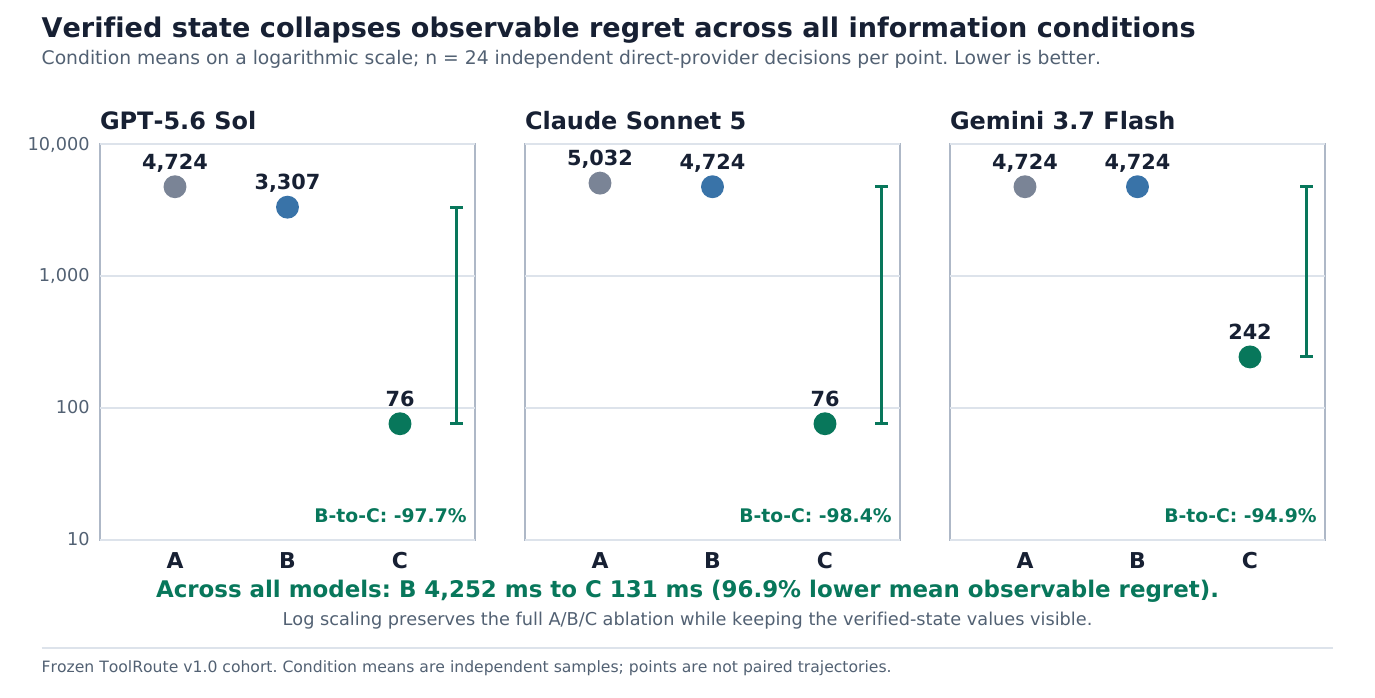}
\caption{Mean observable policy regret for all three information conditions in the frozen 216-decision cohort. Log scaling keeps verified-state values visible; plot labels round values to whole milliseconds. A, B, and C are independent condition means, not trajectories.}
\label{fig:figure1}
\end{figure}

The effect appears in every model family. GPT-5.6 Sol, Claude Sonnet 5, and Gemini 3.7 Flash each complete 23/24 C decisions; their C mean regrets are 75.50\,ms, 75.50\,ms, and 242.35\,ms. More specifically, whenever a C model selected an active route, all \textbf{69/69} selections matched an observable-best route and incurred 0\,ms regret. The three remaining C decisions---one from each model---selected \texttt{wait} when both routes were degraded. They remain in the primary denominator, fail the frozen retrieval task, and account for all residual C regret.

\subsection{Study 2: live transport replication}
\enlargethispage{1.5\baselineskip}

The B-to-C comparison again isolates the state values within the same interface structure. B completes 5/9 live actions (55.6\%; Wilson 95\% CI 26.7--81.1\%) with 8,956.37\,ms mean regret. C selects the observable-best route in \textbf{9/9 cases}, completes \textbf{9/9 live actions} (100.0\%; Wilson 95\% CI 70.1--100.0\%), and incurs \textbf{0.00\,ms} regret. Task-only A completes 6/9 actions (66.7\%; Wilson 95\% CI 35.4--87.9\%) with 6,734.30\,ms regret. Study 2 is an end-to-end mechanism replication across a predeclared model-by-fault matrix, not a standalone population-level significance claim.

The C result holds for all three model families and all three fault regimes. Each model receives one latency, one connection-error, and one HTTP-error case; every C decision selects the viable route from the OpenTelemetry-derived state, and the subsequent proxied HTTP action succeeds.

\begin{table}[htbp]
\centering
\small
\caption{Study 2 live transport outcomes by model family. Regret is mean observable policy regret in milliseconds.}
\label{tab:study2_results}
\begin{tabularx}{\linewidth}{@{} l *{6}{>{\raggedleft\arraybackslash}X} @{}}
\toprule
\textbf{Model} & \textbf{A: succ/3} & \textbf{A regret} & \textbf{B: succ/3} & \textbf{B regret} & \textbf{C: succ/3} & \textbf{C regret} \\
\midrule
GPT-5.6 Sol      & 2   & 6,767.64 & 1   & 13,434.67 & 3   & 0.00 \\
Claude Sonnet 5  & 2   & 6,668.05 & 2   & 6,667.18  & 3   & 0.00 \\
Gemini 3.7 Flash & 2   & 6,767.22 & 2   & 6,767.25  & 3   & 0.00 \\
\midrule
\textbf{Total}   & \textbf{6/9} & \textbf{6,734.30} & \textbf{5/9} & \textbf{8,956.37} & \textbf{9/9} & \textbf{0.00} \\
\bottomrule
\end{tabularx}
\end{table}

\begin{figure}[tp]
\centering
\includegraphics[width=0.93\textwidth]{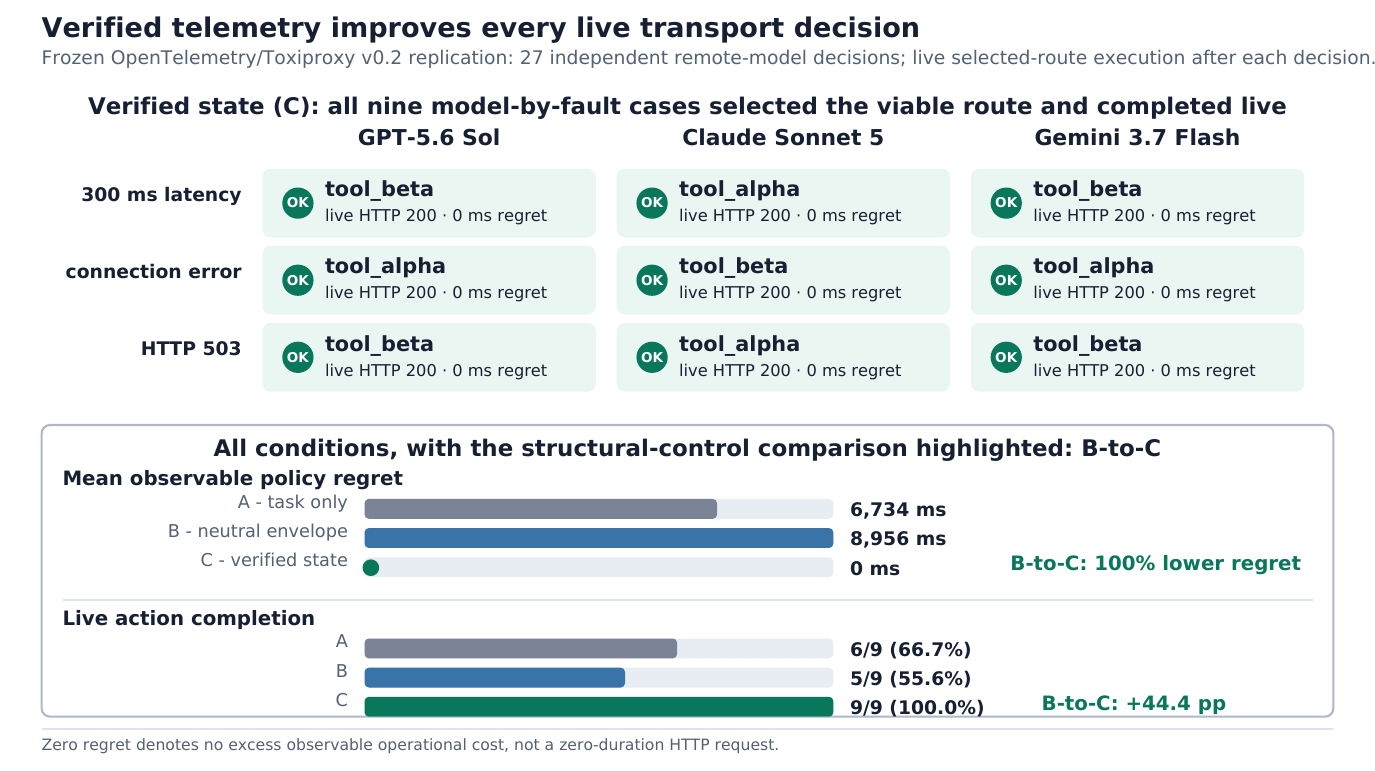}
\caption{Study 2 live transport replication. The upper panel shows that all nine verified-state C cases selected the observable-best route and completed the subsequent live proxied HTTP action. The lower panel reports mean observable policy regret and live completion for A/B/C, highlighting the B-to-C structural comparison. Here, 0\,ms regret denotes zero excess observable operational cost, not zero HTTP duration; ``pp'' denotes percentage points.}
\label{fig:figure2}
\end{figure}

The exact fault and option-order allocation is retained in the frozen manifest. The task template names \texttt{tool\_alpha} before \texttt{tool\_beta}, and the B/C telemetry rows use the same canonical alphabetical order. These regularities plausibly contribute to the observed baseline preference but were not independently manipulated. Across A and B, models selected \texttt{tool\_alpha} in 17/18 decisions (94.4\%), including 7/8 cases in which \texttt{tool\_beta} was listed first. We therefore report a cohort-specific \textbf{default route-label preference}, not a universal positional-bias claim. Condition C removed its operational consequence: every model selected the healthy route regardless of which route was degraded or listed first.

\section{Implications, boundaries, and next tests}
\enlargethispage{\baselineskip}

\subsection{What the studies establish}

The two studies establish a behavioral information effect for the tested mechanism. When task semantics and candidate eligibility were fixed, current verified tool-health state changed pre-dispatch choices and sharply reduced avoidable operational cost relative to both the task-only baseline and the same-shape neutral control. All 69 active C choices in Study 1 and all nine C choices in the live replication selected an observable-best route. The primary study uses a seeded host-owned monitor; the replication shows that the intervention also operates when standard OpenTelemetry spans are reduced from a live local HTTP path.

\subsection{Where the interface belongs}

Harness Awareness complements deterministic recovery rather than replacing it. Proxies, circuit breakers, schedulers, and client libraries should continue to own routine failover, hard safety, and fast recovery. The agent-facing interface becomes relevant when current state changes a task-aware decision: whether to switch to a semantically different provider or source, defer or verify an action, accept a lower-fidelity mode, or restructure a plan under a deadline or quota.

The frozen no-retry protocol isolates first-dispatch discernment before reactive recovery can hide the initial choice. Production retries are effective for many transient faults, but established guidance calls for bounded attempts, backoff and jitter, and idempotency-aware handling because retries can add latency, amplify load, or duplicate side effects~\cite{aws2023retry}. This read-only study does not compare recovery policies or establish benefits for non-idempotent operations. Its natural follow-on is an equal-budget comparison between retry-only recovery and verified-state-first recovery using the same bounded retry policy.

\subsection{Next tests}

Three experiments would extend the demonstrated mechanism most directly. First, \textbf{telemetry-guided recovery} should compare retry-only behavior with verified-state-first behavior under identical error feedback, action budgets, and retry rules. Second, \textbf{freshness-aware dispatch} should vary the delay between observation and action and test host-side revalidation when a snapshot expires. Third, \textbf{heterogeneous operational selection} should present semantically eligible candidates whose capability, answer quality or freshness, price, quota, privacy, and latency trade-offs must be weighted against the task.

The broader SST program also includes runtime, hardware, and economics, but this paper does not treat them as ToolRoute evidence. BATS already establishes the value of explicit tool-call budget awareness~\cite{liu2025bats}, while the separate Substrate Awareness study provides controlled evidence that disclosing RAM and time constraints changes generated programs~\cite{agrawal2026substrate}. Larger tool catalogs and persistent trajectories additionally require compact state deltas, semantic candidate retrieval followed by selective telemetry hydration, and explicit context-serialization budgets.

\section{Conclusion}

Semantic discovery determines what an agent can do; current operational state can still determine how it should do it. ToolRoute shows that, with eligibility held fixed, verified tool-health state produces a large and consistent improvement in pre-dispatch selection under a structural control. A separate OpenTelemetry/Toxiproxy study replicates the mechanism across a live local HTTP path. These results establish selected host telemetry as a consequential decision input for the tested tools-and-network setting. The next questions are how that input composes with reactive recovery, survives state decay, and guides strategic choices among heterogeneous candidates---while the host remains authoritative over truth, enforcement, and execution.

\section{Artifact availability}
\enlargethispage{2\baselineskip}

The complete evaluation archive is available at \url{https://github.com/manu2/scac-dynamic-agent-harness}~\cite{agrawal2026scac}. It preserves the frozen manifests, complete retained study directories, sanitized provider requests and responses, model-visible prompts, host observations, raw OpenTelemetry spans, reducer outputs, action results, finalization hashes, derived analyses, figures, and PDF-build scripts. The two study cohorts remain separately labeled and are never pooled.

\appendix
\section{Reproducibility record}
{\footnotesize
\setlength{\parskip}{1pt}
The Study 1 task asks the model to retrieve the next required record through functionally equivalent \texttt{tool\_alpha} and \texttt{tool\_beta} routes, return exactly one listed action, and choose from a predeclared rotation of the routes plus \texttt{wait}. Condition A receives that task only. Condition B receives the same fixed telemetry envelope and row structure as C, with identical neutral values for both routes: \texttt{window=6}, \texttt{succ=6}, \texttt{consec\_fail=0}, \texttt{latency\_ewma=1000.0ms}, \texttt{last\_err=NONE}, \texttt{circuit=CLOSED}, and \texttt{age=10ms}. Study 2 uses the identical structure with \texttt{window=3}. Condition C receives the host-rendered checkpoint. The archive records every prompt and option order.

For the route-cost equation in Section~3.1, $p_f=0.5$ when no monitor observation is available. \texttt{wait} has no route cost and is instead scored by the separately frozen rule:
$r_{\mathrm{obs}}(\texttt{wait},z)=500+10{,}000-\min_{r\in\mathcal{R}}c_{\mathrm{obs}}(r,z)$~ms.
This represents a 500\,ms deferral and a 10,000\,ms missed-record penalty relative to the best route; it is neither free nor a hidden retry. In the frozen schedule, $\max_z\min_{r\in\mathcal{R}}c_{\mathrm{obs}}(r,z)=8{,}687.95$\,ms, below 10,500\,ms, so this score is non-negative in every evaluated episode.

Study 1 and Study 2 use \texttt{gpt-5.6-sol}, \texttt{claude-sonnet-5}, and \texttt{gemini-3.7-flash}, a 1,024-token output limit, no provider-native tools, and no request retries. OpenAI and Anthropic requests omit an explicit temperature field; Gemini requests set temperature to 0.0. For Study 2, the frozen manifest fixes all 27 model, condition, faulted-route, option-order, and execution-order cells. Each monitor phase performs three OpenTelemetry-instrumented requests per route; the reducer uses $\alpha=0.2$, opens a circuit after three consecutive failures, marks other windows with failure fraction at least 0.4 as \texttt{DEGRADED}, and applies a 2\,s freshness bound. The selected route then executes over the still-live local proxy.

The repository analysis validates the frozen 216-cell Study 1 grid and 27 finalized Study 2 episodes before emitting derived results. Outputs are checked against retained finalization hashes; analysis makes no provider calls and does not rewrite experiment artifacts~\cite{agrawal2026scac}.
}

\end{document}